\documentclass[12pt]{article}

\usepackage{amsmath,amsthm, amsfonts, amssymb, amsxtra, amsopn}
\usepackage{pgfplots}
\usetikzlibrary{calc}
\pgfplotsset{compat=1.13}
\usepackage{pgfplotstable}
\usepackage{graphicx,grffile}
\usepackage{multirow}
\usepackage{booktabs}
\usepackage{listings}
\usepackage{cmap}
\usepackage{colortbl}
\usepackage{adjustbox}
\usepackage{epsfig}

\usepackage[tableposition=top,font=small,skip=5pt]{caption}

\usepackage[explicit]{titlesec}

\PassOptionsToPackage{hyphens}{url}

\usepackage{hyperref}
\hypersetup{colorlinks=true,linkcolor=black,citecolor=black,urlcolor=blue,filecolor=black}
\hypersetup{pdfpagemode=UseNone,pdfstartview=}

\definecolor{darkgreen}{rgb}{0.125,0.5,0.169}

\usepackage{enumitem}
\setlist[itemize]{noitemsep, topsep=0pt}

\pgfplotsset{
  nodes near coords black white/.style={
    small value/.style={
      yshift=-9pt,
      text=black,
      /pgf/number format/fixed,
      /pgf/number format/precision=0,
      /pgf/number format/zerofill=true,
      scale=1.125,
    },
    large value/.style={
      yshift=-9pt,
      text=white,
      /pgf/number format/fixed,
      /pgf/number format/precision=0,
      /pgf/number format/zerofill=true,
      scale=1.125,
    },
    every node near coord/.style={
      check for zero/.code={
        \pgfmathfloatifflags{\pgfplotspointmeta}{0}{
          \pgfkeys{/tikz/coordinate}
        }{
          \begingroup
            \pgfkeys{/pgf/fpu}
            \pgfmathparse{\pgfplotspointmeta<#1}
            \global\let\result=\pgfmathresult
          \endgroup
          \pgfmathfloatcreate{1}{1.0}{0}
          \let\ONE=\pgfmathresult
          \ifx\result\ONE
            \pgfkeysalso{/pgfplots/small value}
          \else
            \pgfkeysalso{/pgfplots/large value}
          \fi
        }
      },
      check for zero,
    },
  },
  nodes near coords black white/.default=500
}

\advance\oddsidemargin by -0.45in
\advance\textwidth by 0.9in

\advance\topmargin by -0.45in
\advance\textheight by 0.9in

\long\def\symbolfootnotetext[#1]#2{\begingroup%
\def\thefootnote{\fnsymbol{footnote}}\footnotetext[#1]{#2}\endgroup}

\newcommand\dunderline[3][-1pt]{{%
  \sbox0{#3}%
  \ooalign{\copy0\cr\rule[\dimexpr#1-#2\relax]{\wd0}{#2}}}}
\def\uuu{\kern-1pt\dunderline{0.75pt}{\phantom{M}}}

\def\zz{\phantom{0}}

\title{AGRO: An Autonomous AI Rover for Precision Agriculture}

\author{Simar Ghumman\footnotemark[1]\,\,\footnotemark[2]\ \ \ 
Fabio Di Troia\footnotemark[1]\ \ \ 
William Andreopoulos\footnotemark[1]\ \ \ \\
Mark Stamp\footnotemark[1]\ \ \ 
Sanjit Rai\footnotemark[3]}

\begin{document}

\symbolfootnotetext[1]{Department of Computer Science, San Jose State University}
\symbolfootnotetext[2]{ghumman.simar$@$gmail.com}
\symbolfootnotetext[3]{Department of Economics, San Jose State University}

\maketitle

\abstract
Unmanned Ground Vehicles (UGVs) are emerging as a crucial tool in the world of precision agriculture. The combination of UGVs with machine learning allows us to find solutions for a range of complex agricultural problems. This research focuses on developing a UGV capable of autonomously traversing agricultural fields and capturing data. The project, known as AGRO (Autonomous Ground Rover Observer) leverages machine learning, computer vision and other sensor technologies. AGRO uses its capabilities to determine pistachio yields, performing self-localization and real-time environmental mapping while avoiding obstacles. The main objective of this research work is to automate resource-consuming operations so that AGRO can support farmers in making data-driven decisions. Furthermore, AGRO provides a foundation for advanced machine learning techniques as it captures the world around it.


\maketitle

\section{Introduction}\label{sec:Introduction}

As the global population continues to increase, the need to maximize food production and minimize resource consumption comes into focus~\cite{daszkiewicz2022}. Precision agriculture is an answer to both. It is a farming strategy that focuses on improving crop yields by monitoring, assessing, and reacting to time-related and location-specific variations~\cite{karunathilake2023}. This strategy focuses on gathering vast amounts of data to make calculated decisions that optimize the available resources. By leveraging technology, farmers can then monitor their farms and make data-driven decisions. For example, in agriculture, using ML we can analyze an image dataset of leaves and learn which crops are diseased~\cite{chen2023_icist22}. This analysis provides information that identifies early signs of plant disease, nutrient deficiencies, and pest activities, offering actionable real-time information to farmers while saving money and time. By streamlining operations, this integration enables farmers to allocate their time and resources more efficiently.

\begin{figure}[!htb]
    \centering
    \includegraphics[scale=0.3]{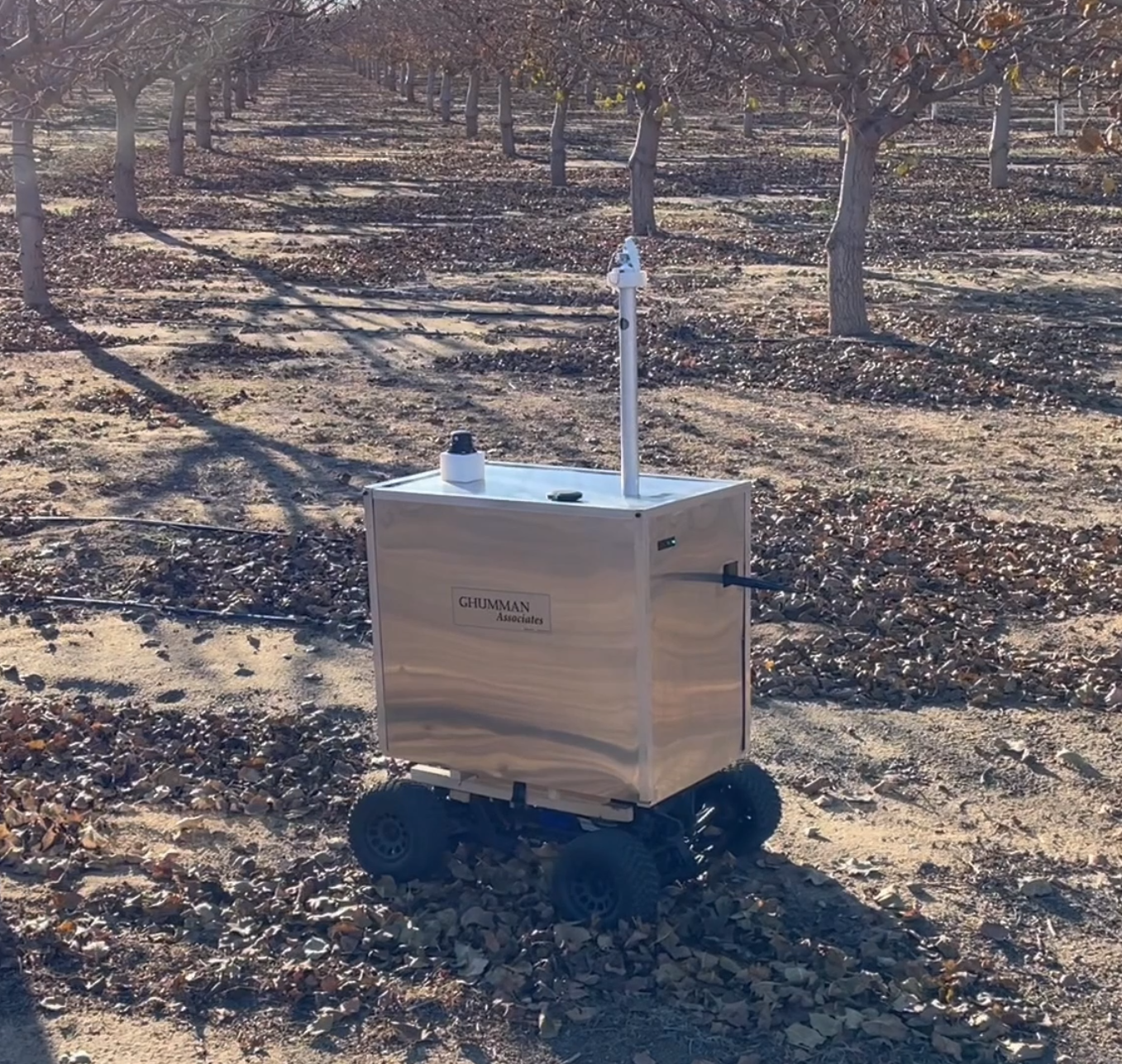}
    \caption{AGRO: An Autonomous AI Rover for Precision Agriculture}
    \label{fig:AGRO}
\end{figure}

The goal of this research is to develop an autonomous robotic system for the agricultural sector, as shown in {Figure~\ref{fig:AGRO}}. This robot is capable of navigating predetermined routes within an agricultural field using Light Detection and Ranging (LiDAR). The robot implements pathing algorithms such as Dijkstra's algorithm and BendyRuler to autonomously traverse farmland and perform object avoidance~\cite{Dijkstra1959, BendyRuler}. Using GPS and LiDAR, the AGRO robot can operate in difficult terrain while avoiding obstacles and dynamically adjusting its path.

To enhance AGRO’s capabilities, we conduct several machine learning experiments using YOLO, an extremely lightweight and efficient model that is used for on-the-go object detection. We train a YOLO-based object detection model to identify and count any pistachios in the field, using a custom-annotated dataset. We also perform hyperparameter tuning using grid search and explore various data augmentation techniques.

The rest of the paper follows the following format. In \hyperref[sec:Background]{Section~\ref{sec:Background}}, we discuss relevant background information and the progress in precision agriculture, as well as papers similar to ours. In \hyperref[sec:Methodology and Equipment]{Section~\ref{sec:Methodology and Equipment}}, we describe our methodology and the equipment used. In \hyperref[sec:Implementation]{Section~\ref{sec:Implementation}}, we outline our implementation. In \hyperref[sec:Experiments]{Section~\ref{sec:Experiments}}, we dive into our experiments and results. Finally, we conclude with \hyperref[sec:Conclusion and Future Work]{Section~\ref{sec:Conclusion and Future Work}}, discussing ideas for future work.

\section{Background}\label{sec:Background}

In this section, we examine representative examples of advancements made through ML and AI techniques in the world of precision agriculture.

\subsection{Advancements in Precision Agriculture}

The world of precision agriculture has experienced significant improvement during the last few years because of AI integration, with a few examples discussed ahead. Research indicates that convolutional neural networks (CNNs) effectively identify plant leaf diseases which leads to high  accuracy in classification~\cite{daszkiewicz2022}. The combination of unmanned aerial vehicles (UAVs) with machine learning algorithms allows farmers to monitor their land and obtain valuable information about crop health and  resource management~\cite{zualkernan2023}. Ensemble learning methods help estimate crop health  by measuring maize leaf area index (LAI) which enhances model accuracy across different environmental conditions according to~\cite{cheng2022}. The implementation of Faster R-CNN deep learning architecture along with other  methods represents recent advancements in plant disease detection which leads to better agricultural results~\cite{abdullahi2022}. AI continues to integrate with ongoing agricultural developments which creates a path toward sustainable farming practices.

\subsection{Yield Estimation Methods}

Another area of research is yield estimation, which primarily relies on two approaches: region-based methods and counting-based methods~\cite{rahnemoonfar2017}. Region-based methods estimate the yield on specific areas of an image by extracting features such as color, texture, or shape from regions that contain fruits or crops. Counting-based methods, on the other hand, first detect objects in an image and then count the total number of objects to  estimate the yield. Both methods have their own advantages and disadvantages, but deep learning combines both of these approaches.

\subsection{Deep Learning Approaches}

Research findings show that deep learning techniques address the challenges of fruit counting such as illumination variance, occlusion by foliage, overlapping fruits, fruits under shadow and scale variation. For example, Chen et al. used convolutional neural networks (CNNs) and fully convolutional networks (FCNs) to create a yield estimation pipeline which effectively counted fruit in unstructured environments~\cite{chen2017_counting}.

In this approach, FCNs are used for semantic segmentation to identify regions likely to contain fruit. This is followed by blob detection and a secondary CNN that analyzes each segmented region and estimates the fruit count. This method achieves over 20\% greater accuracy than previous approaches.

Bargoti and Underwood take a similar deep-learning-based approach, employing multilayer perceptrons (MLPs) and CNNs to perform image segmentation and subsequently count the fruits~\cite{bargoti2017}. These methods show that deep learning can learn the relevant features to perform accurate detection and counting even under challenging conditions such as occlusion and variable lighting.

\subsection{Our Approach: Yield Estimation with YOLO}

For our approach, we use a neural network called You Only Look Once (YOLO), similar to the work of Osman et al., but using a ground-based vehicle without Deep Simple Online and Realtime Tracker (DeepSORT)~\cite{osman2021}. 

YOLO is a real-time object detection algorithm. It treats detection as a regression problem rather than a classification task by separating boundary boxes and associating probabilities to each detected image using a single convolutional neural network. In our approach, YOLO is used to locate each pistachio within the field and count the detected instances. The robot captures images from approximately 10 feet away while navigating along the center of the row between pistachio trees.

DeepSORT is an object tracking algorithm that builds upon SORT (Simple Online and Realtime Tracker) by integrating deep learning-based appearance descriptors. It is commonly used in vision-based applications to track objects across multiple frames. However, since our system focuses solely on one-time detection rather than continuous tracking, we do not integrate DeepSORT into our approach.

YOLO is widely used in the agriculture space for its exceptional speed in real-time object detection, making it an ideal choice for our application.

\section{Methodology and Equipment}\label{sec:Methodology and Equipment}

In this section, we describe the steps taken to build and optimize AGRO.

\subsection{Prototype Development}

\begin{figure}[!htb]
    \centering
    \includegraphics[scale=0.14]{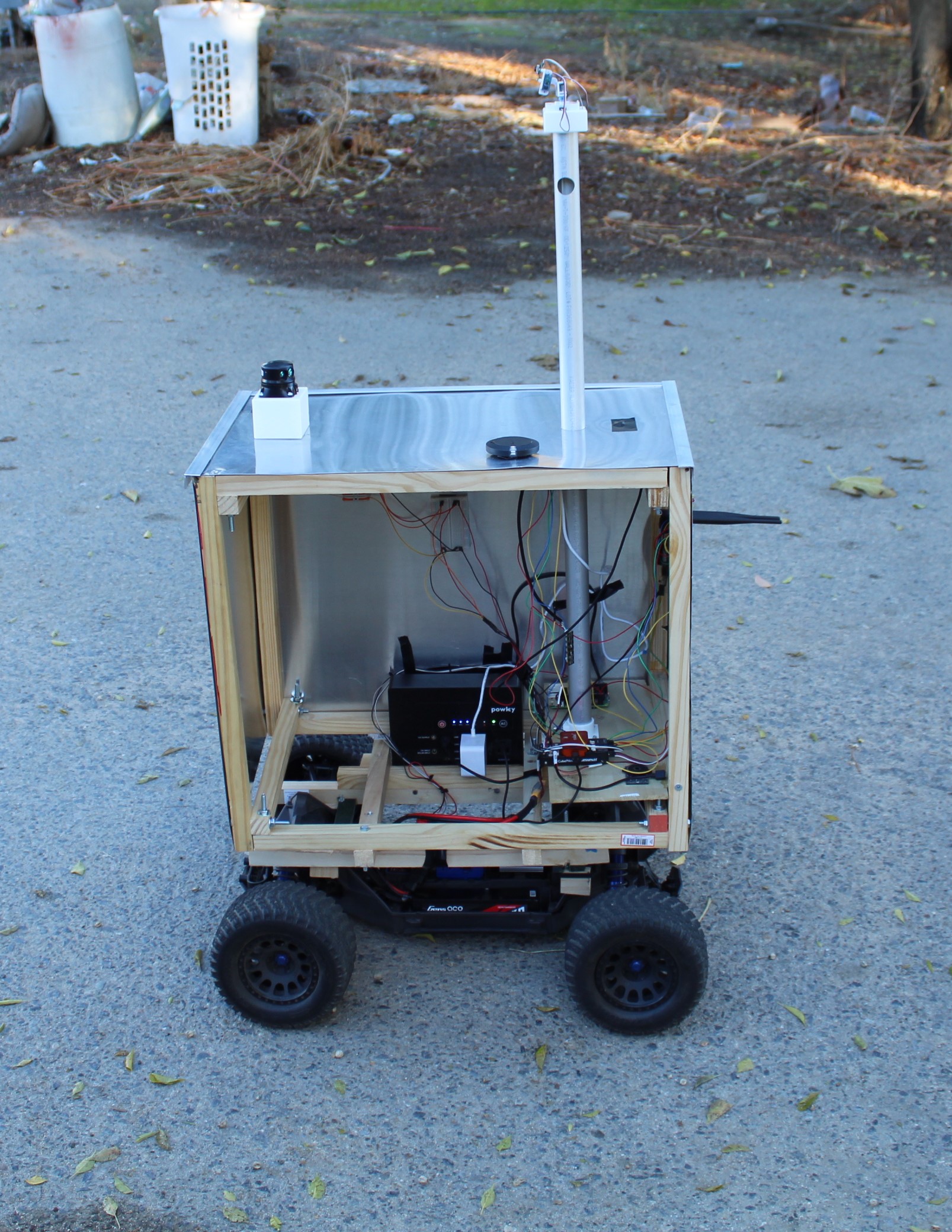}
    \caption{AGRO: Interior system}
    \label{fig:Inside_Look}
\end{figure}

As shown in {Figure~\ref{fig:Inside_Look}}, the development of the AGRO frame starts from the ground up, with the Traxxas XRT 8S Brushless Electric Race Truck serving as the base platform~\cite{TraxxasXRT8S}. A custom robotic chassis is then used to cover and protect the more fragile components added, preventing damage in the field. The main hardware components used in this research are the following:

\begin{enumerate}
    \item Cube Orange+ Standard Set ADS-B: Used as the motherboard to control all other components~\cite{CubeOrange}.
    \item Here4 Multiband RTK GNSS: Provides accurate Global Positioning System (GPS) coordinates and functions as a compass for navigation routes~\cite{Here4RTK}.
    \item LightWare SF45/B 350 LiDAR: Used for object avoidance, enabling autonomous driving~\cite{LightWareLiDAR}.
    \item RFD900x Telemetry Bundle: Used for communication with the rover from afar~\cite{RFD900x}.
    \item Raspberry Pi 5: Connected to the Cube Orange, it reads the GPS coordinates and captures images of the route with the Arducam 64MP Hawkeye camera~\cite{RaspberryPi5}.
    \item Arducam 64MP Hawkeye Camera: Captures high-resolution images to assist in data collection~\cite{Arducam64MP}.
    \item Power Supply: Powered by LiPo batteries and a mounted power bank~\cite{LiPoBattery}.
    \item FrSky Taranis Receiver X8R 8 Channel 2.4GHz ACCST \& SBUS: Connected to the transmitter for control of the rover~\cite{Frsky}.
    \item FrSky Taranis X9D Plus 2019 Transmitter: Connected to the receiver to serve as a controller for the rover~\cite{Frsky}.
\end{enumerate}

We control the AGRO through a software called, Mission Planner. This software allows for remote control, telemetry data monitoring, configurations, and mission planning~\cite{MissionPlanner}.

\subsection{Camera System Exploration}

When selecting the appropriate camera for AGRO's data capture system, we choose to focus on RGB imaging, as it is capable of detecting changes in color that help reveal differences in the crop and is cost-efficient. We also experiment with different capture resolutions and RGB cameras, and eventually choose the Arducam 64MP Hawkeye camera for its high-resolution images.

\subsection{Autonomous Navigation Techniques}

Autonomous navigation is a crucial aspect for the rover. For this process, we use the LightWare SF45/B LiDAR for obstacle avoidance, employing an algorithm that combines two different pathfinding approaches, Dijkstra's algorithm and BendyRuler. Dijkstra's algorithm determines the shortest path to a waypoint, while BendyRuler enables AGRO to dynamically navigate around obstacles. This combination allows AGRO to autonomously navigate through pre-defined routes using the Here4 GPS for directions.

\section{Implementation}\label{sec:Implementation}

In this section, we discuss the implementation on both the hardware and software components.

\subsection{Hardware Assembly and Integration}

AGRO is implemented in a series of well-defined stages. The first stage involves assembling the robotic chassis and incorporating the Cube Orange as the main processing unit. We then integrate the necessary sensors such as LiDAR and GPS. For the robot chassis, the Traxxas XRT 8S Brushless Electric Race Truck is chosen, as it comes with all the components pre-assembled and tested.

The Electronic Speed Controller (ESC), motors, and servos are then connected to the Cube Orange, allowing movement through the Mission Planner software. Afterwards, the FrSky Taranis Receiver is connected to the Cube Orange and bound to the FrSky Taranis Transmitter, allowing manual control of the rover. The sticks and switches on the transmitter are then mapped as Pulse Width Modulation (PWM) signals to the receiver, and Mission Planner is used to assign functionality to each control.

\subsection{Mission Planner Setup}

After assembling the hardware, the software phase involves setting up Mission Planner to handle autonomous navigation tasks. In Mission Planner before activating the rover, the system performs multiple checks to verify proper operation, such as:

\begin{itemize} 
    \item Calibration of accelerometer, gyroscope, compass, and the Attitude and Heading Reference System (AHRS). 
    \item GPS status check for the current location. 
    \item Verification of RC signal integrity and throttle neutrality. 
    \item Confirmation of failsafe settings, battery voltage, Extended Kalman Filter (EKF) health, vibration levels, and internal hardware status. 
    \item Proper logging setup and verification that tuning parameters are appropriate. 
\end{itemize}

\begin{figure}[!htb]
\centering
\includegraphics[scale=0.125]{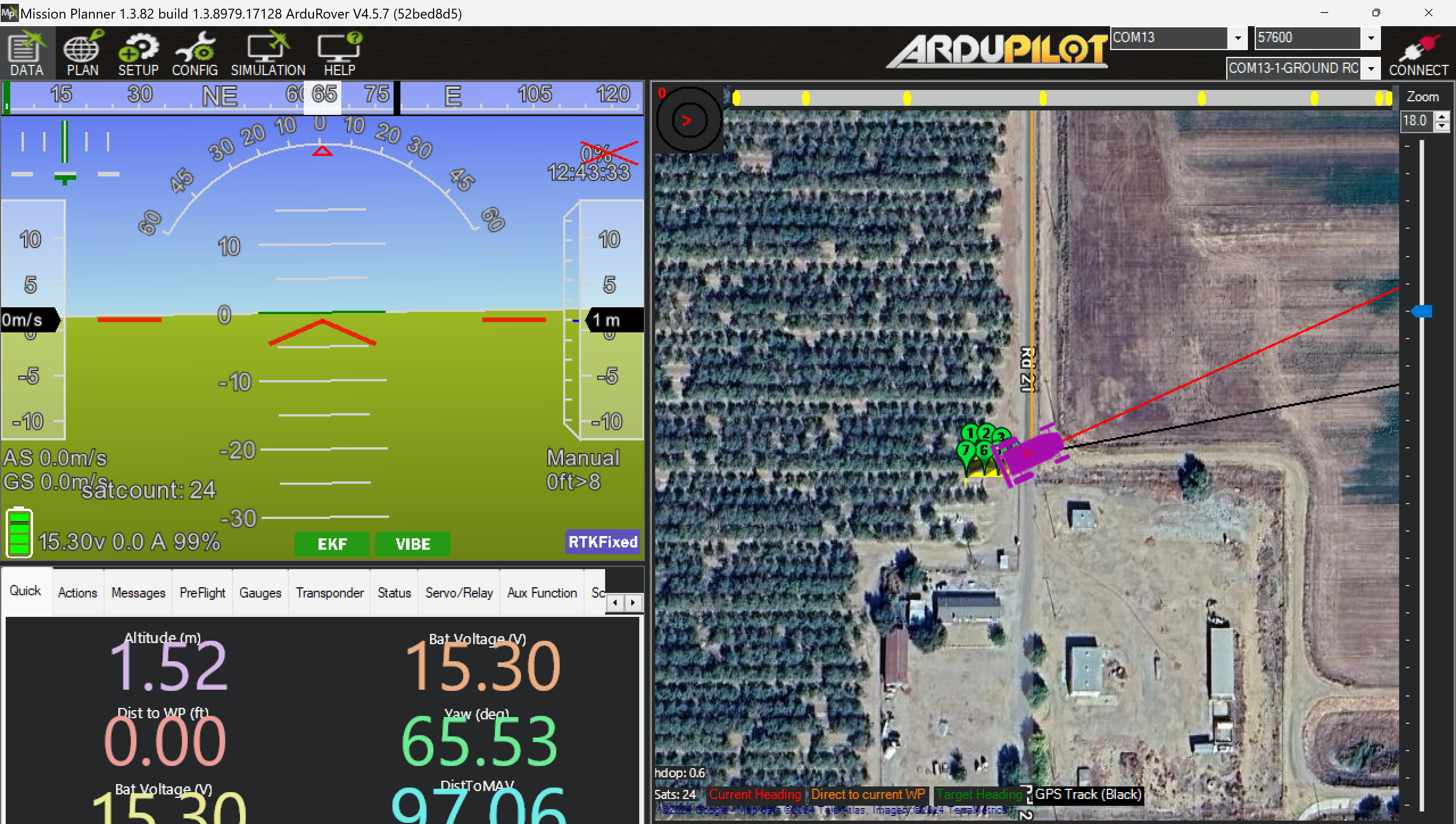}
\caption{Mission Planner Software: Planning out mission}
\label{fig:mission_planner}
\end{figure}

Once the pre-arm checks pass safely, we arm the rover, which allows us to flash our custom mission waypoints to the rover through Mission Planner. We can then track the rover in real-time as it completes its mission, manually issue commands to control the rover, and adjust the mission parameters if needed. All metrics, such as ground speed, distance to waypoint, and altitude, are monitored in real-time through Mission Planner as shown in {Figure~\ref{fig:mission_planner}}.

We also use Real Time Kinematics (RTK), a type of GPS technology that provides centimeter-level accuracy. RTK improves positioning accuracy by using a stationary base station at a known location, which compares its position to that of a mobile receiver, in our case the rover. The system calculates errors caused by atmospheric delays, satellite clock errors, and signal noise. These corrections are transmitted to the rover in real-time, typically via a radio link or Internet connection, depending on the setup. 

In contrast, standard GPS determines position based only on satellite signals, resulting in positioning errors of several meters. This imprecision causes the rover to have difficulties accurately traversing the farmland. However, RTK resolves these issues by providing real-time corrections for precise positioning.

To integrate RTK, we use the Here4 GPS device, which supports both standard GPS and RTK modes. We connect to a nearby local base station using the Networked Transport of Radio Technical Commission for Maritime Services (RTCM) via Internet Protocol to receive corrections. Point One Navigation offers these services via an Networked Transport of RTCM via Internet Protocol (NTRIP) server. By connecting to it, we were able to achieve an RTK Fixed or RTK Float connection, significantly improving our rover's sense of direction~\cite{PointOneNavigation}.

\subsection{Camera Integration and Data Collection}

For the implementation of the camera, we use a Raspberry Pi connected to an Arducam 64MP Hawkeye camera to record the entire mission on the rover. A boot-up script operating on the Raspberry Pi establishes a connection to the Cube Orange flight controller through a Universal Serial Bus (USB) to retrieve GPS coordinates.

The script operates by sending a high signal from the Cube Orange to the Raspberry Pi when the designated waypoint is reached which triggers the camera to take a photo. The system captures images at important points of interest while continuously obtaining GPS coordinates through the USB connection.

The system uses three  General-Purpose Input/Output (GPIO) LEDs for visual state monitoring.

\begin{itemize} 
    \item \textbf{Red LED}: Indicates startup and script booting. 
    \item \textbf{Yellow LED}: Shows connection status to the Cube Orange and signals readiness to capture a photo. 
    \item \textbf{Green LED}: Activates during photo capture and GPS data collection. 
\end{itemize}

\begin{figure}[!htb]
\centering
\includegraphics[scale=0.45]{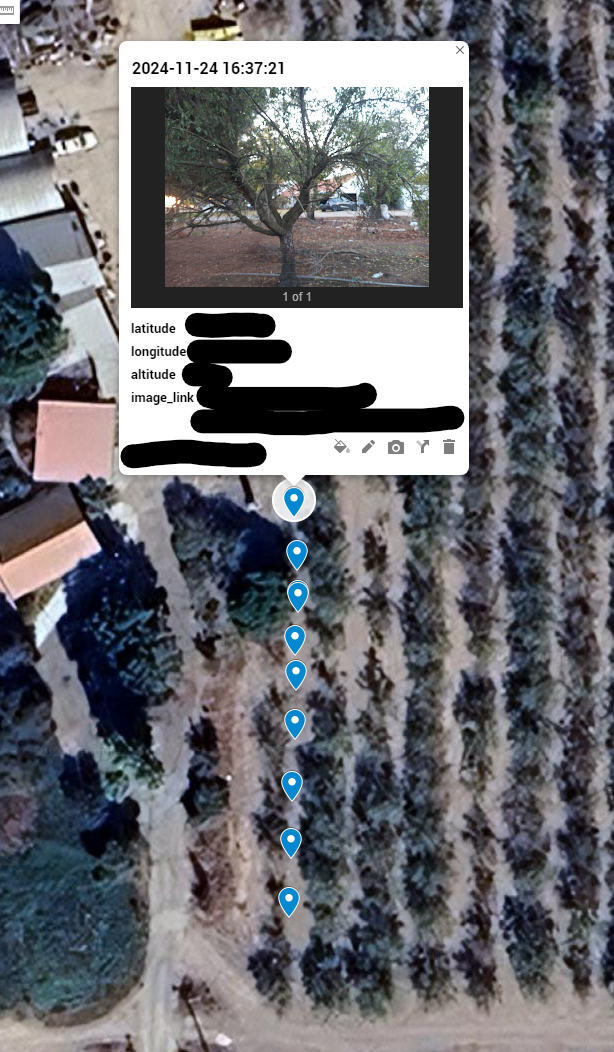}
\caption{Google My Maps visualization of farm mission}
\label{fig:google_mymaps}
\end{figure}

Once the mission is completed, we remove the SD card from the Raspberry Pi extracting the images and corresponding GPS coordinates.  The images are then uploaded to Google Maps: My Maps service, which enables us to geotag each photo to its GPS coordinates as shown in  {Figure~\ref{fig:google_mymaps}}.

\section{Experiments}\label{sec:Experiments}

In this section, we examine the dataset creation, model training, and evaluation results using YOLO.

\subsection{Dataset Creation} 

To train AGRO for object detection, we begin by creating a custom dataset, as publicly available datasets with annotations for pistachios are lacking. Emphasizing high-quality inputs, images are captured at 9152 x 6944 dimensions (64MP), at different points in the field and from various angles. A web application known as Roboflow is used to annotate the dataset~\cite{Roboflow}. However, the site imposes a restriction that only allows a maximum file size of 12MP. To compensate for the limited upload size, a Python script is written to split the 64MP image into 6 equal parts. 

\begin{figure}[!htb]
\centering
\includegraphics[scale=0.4]{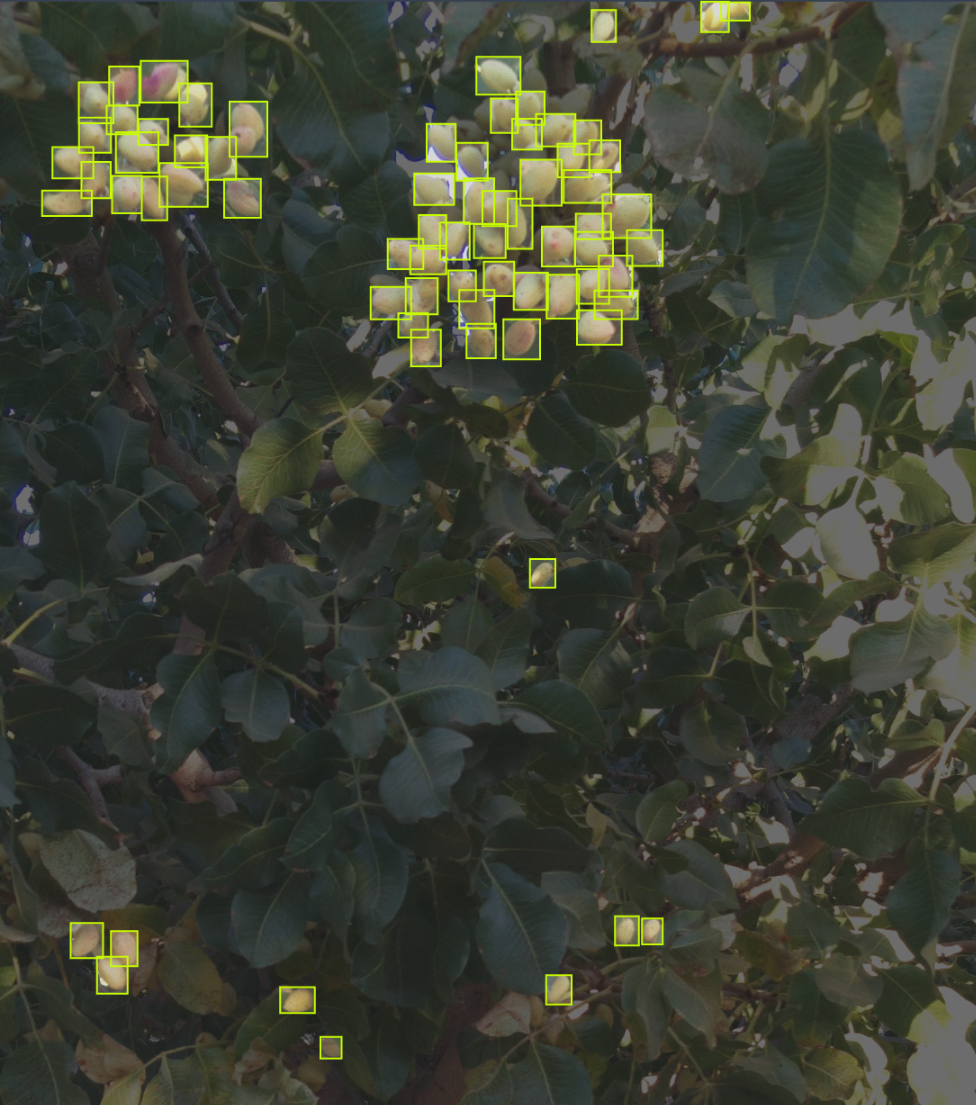}
\caption{Roboflow annotation example: 71 pistachios annotated}
\label{fig:Pistachios_Annoated}
\end{figure}

Once uploaded, annotations are done manually by drawing boundary boxes around each visible pistachio. Any occlusion by foliage or blurry photos is \hbox{discarded}. Multiple revisions are performed to limit human error, resulting in a clean dataset of 5399 positive annotations across 145 images. A sample annotated image is shown in {Figure~\ref{fig:Pistachios_Annoated}}.

To further improve model training, we incorporate negative sample images that contain no pistachios. YOLO incorporates these samples through an implicit background learning mechanism, where any region not annotated with an object is considered background. These negative samples are then used to reinforce the model’s understanding that “no detection” is sometimes the correct choice.

For this process, we use the original dataset and crop the largest area without pistachios in each image to create 145 background images, each labeled as negative. We then combine these 145 negative annotations with the original positive annotations. This results in our initial dataset of 5544 annotations across 145 images made up of 5399 positive annotations and 145 negative annotations.

\subsubsection{Data Augmentation} 

For the preprocessing phase, we auto-orient each image and then apply multiple data augmentation techniques to enhance the model's robustness. The following augmentations are applied:

\begin{enumerate} 
    \item \textbf{Hue Adjustment:} Randomly adjusted between $\hbox{}-\!15^{\circ}$ and $\hbox{}+\!15^{\circ}$. 
    \item \textbf{Brightness Adjustment:} Randomly modified between $\hbox{}-\!15\%$ and $\hbox{}+\!15\%$ to simulate varying lighting conditions. 
    \item \textbf{Exposure Adjustment:} Randomly altered between $\hbox{}-\!10\%$ and $\hbox{}+\!10\%$. 
    \item \textbf{Blur:} Applied up to 2.5px to simulate slight defocus or motion blur. 
    \item \textbf{Noise:} Added noise to up to 0.1\% of pixels to help the model become more resilient. 
\end{enumerate}

These augmentations ensure that the YOLO model is capable of detecting pistachios under any conditions. This leads to an augmented dataset size of 20556 annotations across 545 images, containing 20011 positive samples and 545 negative samples.

\subsection{Model Training} 

To begin the model training process, the annotated dataset gets exported to Google Drive and we import the YOLO library through an Nvidia A100 GPU with High-RAM. The model receives a pretrained  YOLOx (extra-large) weight file to establish transfer learning which enhances its performance.

We then perform a grid search to determine hyperparameters over the values presented in Table~\ref{tab:performance_metrics}. However, we have to change the weight file to YOLOn (nano) weight file in order to conduct this training, as the resources in Google Colab are limited. When viewing the best parameters, we are only able to pick the second best, as increasing both image size and batch size produces an unable to allocate memory error on Google Colab.

\begin{table}[!htb] 
\caption{Performance metrics across epochs, image sizes, and batch sizes}
\label{tab:performance_metrics}
\centering
\begin{tabular}{cccc}
\toprule
Epochs & Image Size & Batch Size & mAP@50\\
\midrule
\zz25     & 640        & 16         & 0.8651\\
\zz25     & 640        & 32         & 0.8695\\
\zz25     & 800        & 16         & 0.9060\\
\zz25     & 800        & 32         & 0.9161\\
\zz50     & 640        & 16         & 0.8963\\
\zz50     & 640        & 32         & 0.8945\\
\zz50     & 800        & 16         & 0.9336\\
\zz50     & 800        & 32         & 0.9352\\
100    & 640        & 16         & 0.9242\\
100    & 640        & 32         & 0.9296\\
\textbf{100} & \textbf{800} & \textbf{16} & \textbf{0.9558}\\
\textbf{100} & \textbf{800} & \textbf{32} & \textbf{0.9559}\\
\bottomrule
\end{tabular}
\end{table}

After we select the hyperparameters, the dataset is then split into training, validation, and test sets using an 80:10:10 stratified split at the image level. This preserves the full context of each image, although each image may contain a varying number of pistachio annotations. In addition, to address class imbalance, we took the 545 negative samples initially present (where each image originally had a single negative annotation) and converted them into 545 separate images with empty annotation files. This adjustment yields a final dataset of 1090 images with 20011 positive annotations of which 545 are negative images, allowing the YOLO model to learn background (no detection) effectively.

\subsection{Evaluation and Results} 

To assess the performance of our trained YOLOv10 model, we evaluate it using the following key metrics:

\begin{itemize}
 \item \textbf{Confidence Threshold}: This parameter determines the minimum confidence score required for a detection to be considered valid. The YOLO model uses a default confidence threshold of 0.25, meaning predictions with confidence scores below this threshold are discarded.

 \item \textbf{Intersection over Union (IoU) Threshold}: IoU measures the overlap between the predicted and ground-truth bounding boxes. YOLO's default IoU threshold is 0.70, meaning a detection is valid only if its predicted bounding box overlaps with at least 70\% of the actual object.
 
 \item \textbf{Mean Average Precision at IoU 0.50} (\( \text{mAP}@50 \)): This metric gives the model's average precision when having an IoU of 0.5. This means that a prediction that covers more than 50\% of the ground truth is considered a valid prediction. 

 \item \textbf{Mean Average Precision at IoU 0.50-0.95} (\( \text{mAP}@50-95 \)): This metric provides a more comprehensive evaluation by computing the average precision across multiple IoU thresholds, specifically from 0.5 to 0.95 at intervals of 0.05. A high \text{mAP}@50-95 indicates that the model reliably performs well across various degrees of precision.
\end{itemize}

\begin{figure}[!htb]
\centering
\includegraphics[scale=0.5]{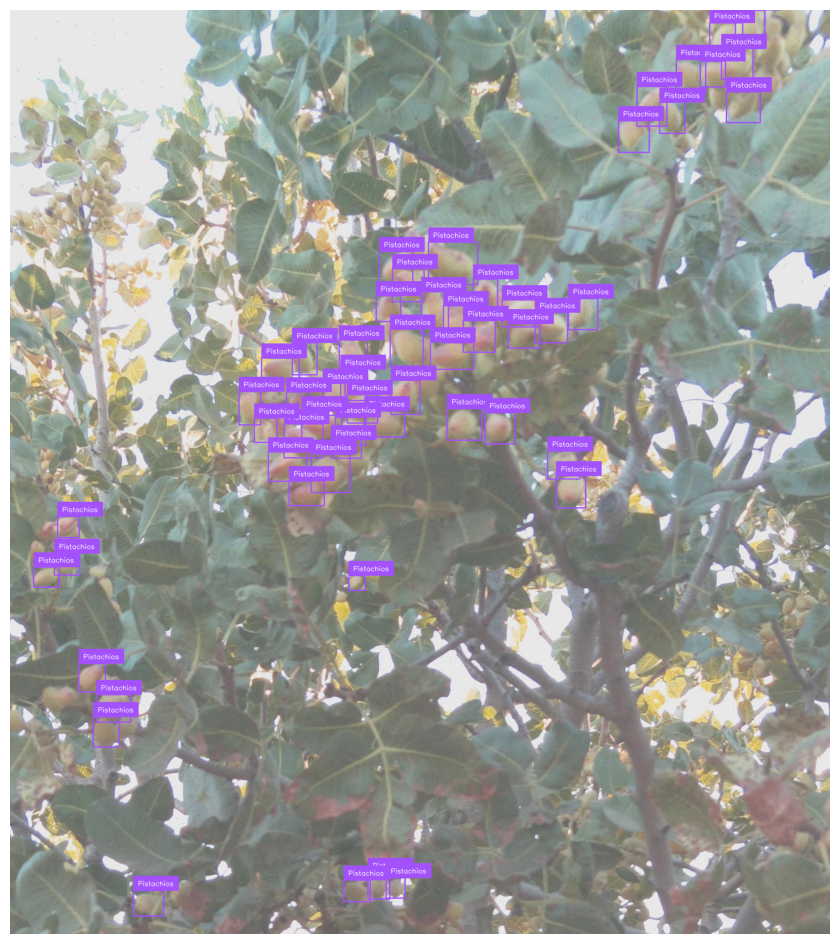}
\caption{YOLOv10 performance on random image}
\label{fig:yolo10_performance}
\end{figure}

By training this model with the finalized hyperparameters and dataset, we achieve a mean average precision (\( \text{mAP}@50 \)) of 0.9888 using YOLOv10. We can also see in {Figure~\ref{fig:Yolo10_Pistachios_Confusion_Matrix}} the confusion matrix for the model, and in {Figure~\ref{fig:yolo10_performance}} you can see how the model performs when detecting the pistachios.

To clarify our evaluation approach, we describe below how we interpret the confusion matrix and compute accuracy.

\subsubsection{Evaluation and the Confusion Matrix} 

In object detection, evaluation is performed at the bounding box level rather than at the image level. For the background images, no bounding boxes are annotated. If the model correctly produces no detections on these images, that behavior is implicitly considered correct. However, because background encompasses an infinite number of potential regions, we do not explicitly count “true negatives” in the confusion matrix. The confusion matrix reflects only the predicted and ground-truth bounding boxes for pistachios. In our single-class detection task, where only pistachios are annotated, the background (negative samples) is used during training and evaluation only to reinforce that no detections should occur.

\subsubsection{Accuracy Calculation}

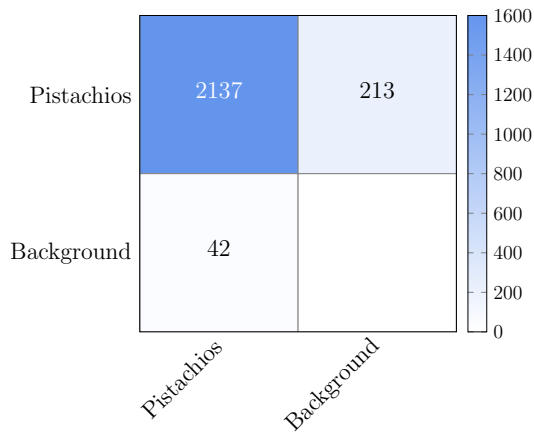
\begin{figure}[!htb]
    \centering
    \begin{tikzpicture}[scale=0.5]
        \begin{axis}[
            width=10cm,
            height=10cm,
            colormap={bluewhite}{color=(white) rgb255=(100,149,237)},
            xticklabels={Pistachios,Background},
            xtick={0,1},
            xtick style={draw=none},
            xticklabel style={anchor=east,rotate=45,yshift=-5pt,scale=1.5},
            yticklabels={Pistachios,Background},
            ytick={0,1},
            ytick style={draw=none},
            yticklabel style={scale=1.5},
            enlargelimits=false,
            colorbar,
            colorbar style={
                ytick={0,200,400,600,800,1000,1200,1400,1600},
                yticklabels={0,200,400,600,800,1000,1200,1400,1600},
                yticklabel={\pgfmathprintnumber\tick},
                yticklabel style={
                    scale=1.25,
                    /pgf/number format/fixed,
                    /pgf/number format/precision=0,
                    /pgf/number format/1000 sep={}
                }
            },
            point meta min=0,
            point meta max=1600,
            nodes near coords={\pgfmathprintnumber\pgfplotspointmeta},
            nodes near coords black white,
            every node near coord/.append style={/pgf/number format/1000 sep={},scale=1.35}
        ]
            \addplot[
                matrix plot,
                mesh/cols=2,
                point meta=explicit,
                draw=gray
            ] table [meta=C] {
                x y C
                0 0 2137
                1 0 213
                0 1 42
                1 1 0
            };
        \end{axis}
    \end{tikzpicture}
    \caption{YOLOv10 confusion matrix}
    \label{fig:Yolo10_Pistachios_Confusion_Matrix}
\end{figure}

Using YOLO's default confidence threshold of 0.25 and an IoU threshold of 0.70 with the YOLOx weight file, we compute the accuracy of the YOLOv10 model on the test subset. Our test set consists of 10\% of randomly selected images from the dataset, each containing a variable number of pistachio annotations. By summing all observations from the confusion matrix, we find that the test subset contains 2392 samples, consisting of 2350 pistachio samples and 42 background samples. This represents approximately 11.95\% of the total annotations, which we deem sufficiently close to our target of 10\% given that the split was generated by randomly selecting images rather than individual pistachios. Using the confusion matrix values as shown in Figure~\ref{fig:Yolo10_Pistachios_Confusion_Matrix}, we calculate the accuracy as
\begin{align*}
\text{Accuracy} &= \frac{\text{True Positives} + \text{True Negatives}}{\text{Total Observations}} \times 100 \\[1.0ex]
	&= \frac{2137 + 0}{2137 + 213 + 42 + 0} \times 100 \\[1.0ex]
	&= 89.34\%
\end{align*}

We also compare the performances between YOLOv10 and YOLOv11 models as shown in Table~\ref{tab:yolo_metrics}.

\begin{table}[!htb] 
\caption{Performance metrics for YOLOv10 and YOLOv11 models}\label{tab:yolo_metrics}
\centering
\begin{tabular}{ccccc}
\toprule
\raisebox{5pt}{Model} 
  & \shortstack{Final\\Precision} & \shortstack{Final\\Recall} & \shortstack{Final\\mAP@50} & \shortstack{Final\\mAP@50-95} \\
\midrule
YOLOv10 & 0.9603 & 0.9555 & 0.9888 & 0.8536 \\
YOLOv11 & 0.9238 & 0.9236 & 0.9744 & 0.7457 \\
\bottomrule
\end{tabular}
\end{table}

The metrics indicate YOLOv10 performs better than YOLOv11 for both object detection and localization tasks because it produces higher precision (96.03\% vs. 92.38\%) and recall (95.55\% vs. 92.36\%). The more stringent IoU thresholds in YOLOv11 might explain this performance difference. Additionally, YOLOv10 achieves a superior mAP@50 of 98.88\% compared to YOLOv11’s 97.44\%. When considering mAP@50-95, which evaluates localization accuracy across different IoU thresholds, YOLOv10 again outperforms YOLOv11, scoring 85.36\% compared to YOLOv11’s 74.57\%. We see that the difference between mAP@50 and mAP@50-95 comes down to accuracy and precise localization, which requires an even stricter policy regarding annotations and data quality.

YOLOv11 demonstrates better performance in handling extensive and intricate datasets which contain high variability in object appearance, scale, and background conditions. YOLOv11's performance may benefit from both growing the dataset size and making it more diverse. Additionally, further fine-tuning may also improve results, to optimize YOLOv11 performance.

\section{Conclusion and Future Work}\label{sec:Conclusion and Future Work}

In this paper, we presented AGRO, an autonomous AI-powered rover designed for precision agriculture. AGRO was equipped with LiDAR, RTK-GPS, and machine learning models to help avoid obstacles, navigate through the farmland, and estimate crop yield. This system incorporated Dijkstra's algorithm and BendyRuler for efficient pathfinding and could be managed through the Mission Planner software.

Furthermore, we trained a YOLOv10 model to identify pistachios in real-world conditions, achieving an accuracy of \textbf{89.34\%}. A custom dataset was created, and our experiments demonstrated that YOLOv10 achieved superior performance in object detection and yield estimation.

Overall, AGRO demonstrated excellent performance in the field and showed potential to be a useful tool for precision agriculture. However, there is always room to improve. Currently, data extraction was performed using an SD card reader. Implementing an onboard computer such as the Jetson Xavier or integrating a cloud-based system would have significantly improved efficiency, enabling farmers to view critical areas in real-time on their devices. Furthermore, incorporating a multiple camera strategy to capture different angles could have provided a more holistic view of the target land.

Another avenue for improvement is the exploration of GAN-based data augmentation techniques. Deep Convolutional Generative Adversarial Networks (DCGANs) could be leveraged to generate synthetic pistachio images to supplement real training data. However, preliminary experiments showed that naive integration of GAN-generated images led to overfitting, indicating the need for higher variability in synthetic data and manual annotation verification. Future work needs to concentrate on enhancing GAN architectures and improving synthetic data quality to guarantee robust model  training.

Future work can also investigate other machine learning architectures such as Faster Region-based Convolutional Neural  Networks, transformer-based detectors like DEtection TRansformer, and other state-of-the-art models.  Further adjustments to hyperparameters, including confidence thresholds and IoU settings, could provide deeper insights and lead  to improved detection accuracy custom tailored to specific agricultural solutions.

Addressing such limitations will help AGRO navigate more accurately and autonomously, making it an even more effective tool for precision agriculture.

\bibliographystyle{plain}
\bibliography{references.bib}

\begin{thebibliography}{10}

\bibitem{abdullahi2022}
H.~S. Abdullahi and R.~E. Sheriff.
\newblock Introduction to deep learning in precision agriculture: farm image
  feature detection using unmanned aerial vehicles through classification and
  optimization process of machine learning with convolution neural network.
\newblock In R.~C. Poonia, V.~Singh, and S.~R. Nayak, editors, {\em Deep
  Learning for Sustainable Agriculture: Cognitive Data Science in Sustainable
  Computing}, pages 81--107. Academic Press, 2022.

\bibitem{Arducam64MP}
Arducam {64MP} {H}awkeye camera.
\newblock \url{https://www.arducam.com/}, 2023.

\bibitem{BendyRuler}
Ardupilot: {BendyRuler} --- {O}bstacle avoidance algorithm.
\newblock
  \url{https://ardupilot.org/copter/docs/common-oa-bendyruler.html#common-oa-bendyruler},
  2023.

\bibitem{bargoti2017}
S.~Bargoti and J.~P. Underwood.
\newblock Image segmentation for fruit detection and yield estimation in apple
  orchards.
\newblock {\em Journal of Field Robotics}, 34(6):1039--1060, 2017.

\bibitem{chen2023_icist22}
S.~Chen, B.~Su, J.~Tang, and H.~Xie.
\newblock Recognition of plant leaf diseases based on convolutional neural
  network.
\newblock In {\em Proceedings of the 4th International Conference on
  Intelligent Science and Technology (ICIST '22)}, pages 8--14, New York, NY,
  USA, 2023. Association for Computing Machinery.

\bibitem{chen2017_counting}
S.~W. Chen et~al.
\newblock Counting apples and oranges with deep learning: A data-driven
  approach.
\newblock {\em IEEE Robotics and Automation Letters}, 2(2):781--788, 2017.

\bibitem{cheng2022}
Q.~Cheng et~al.
\newblock Estimation of maize {LAI} using ensemble learning and uav
  multispectral imagery under different water and fertilizer treatments.
\newblock {\em Agriculture}, 12(8):1267, Aug 2022.

\bibitem{CubeOrange}
{CubePilot}: {C}ube {O}range+ {S}tandard {S}et {ADS-B}.
\newblock \url{https://www.cubepilot.com/}, 2023.

\bibitem{daszkiewicz2022}
T.~Daszkiewicz.
\newblock Food production in the context of global developmental challenges.
\newblock {\em Agriculture}, 12(6):832, Jun 2022.

\bibitem{Dijkstra1959}
E.~W. Dijkstra.
\newblock A note on two problems in connexion with graphs.
\newblock {\em Numerische Mathematik}, 1(1):269--271, 1959.

\bibitem{Roboflow}
B.~Dwyer et~al.
\newblock Roboflow (version 1.0) [software].
\newblock \url{https://roboflow.com}, 2024.

\bibitem{Frsky}
{FrSky} {T}aranis {R}eceiver {X8R} 8 channel {2.4GHz} {ACCST}.
\newblock \url{https://www.frsky-rc.com/}, 2023.

\bibitem{Here4RTK}
{Hex Technology: Here4 Multiband RTK GNSS}.
\newblock \url{https://www.hex.aero/}, 2023.

\bibitem{karunathilake2023}
E.M.B.M. Karunathilake et~al.
\newblock The path to smart farming: Innovations and opportunities in precision
  agriculture.
\newblock {\em Agriculture}, 13(8):1593, Aug 2023.

\bibitem{LightWareLiDAR}
{LightWare SF45/B 350 LiDAR}.
\newblock \url{https://www.lightwarelidar.com/}, 2023.

\bibitem{MissionPlanner}
M.~Oborne.
\newblock Mission planner: Ground control station for {ArduPilot}.
\newblock \url{https://ardupilot.org/planner/}, 2023.

\bibitem{osman2021}
Y.~Osman et~al.
\newblock Yield estimation and visualization solution for precision
  agriculture.
\newblock {\em Sensors}, 21(19):6657, Oct 2021.

\bibitem{PointOneNavigation}
{Point One Navigation}: Precise positioning for autonomous systems.
\newblock \url{https://pointonenav.com/}, 2024.

\bibitem{rahnemoonfar2017}
M.~Rahnemoonfar and C.~Sheppard.
\newblock Deep count: Fruit counting based on deep simulated learning.
\newblock {\em Sensors}, 17(4):905, Apr 2017.

\bibitem{RaspberryPi5}
{Raspberry Pi Foundation}: Raspberry {P}i 5.
\newblock \url{https://www.raspberrypi.com/}, 2023.

\bibitem{RFD900x}
{RFDesign}: {RFD900x Telemetry Bundle}.
\newblock \url{https://store.rfdesign.com.au/}, 2023.

\bibitem{TraxxasXRT8S}
{Traxxas XRT 8S Brushless Electric Race Truck}.
\newblock \url{https://traxxas.com/products/models/electric/xrt}, 2023.

\bibitem{LiPoBattery}
Turnigy {LiPo} battery pack.
\newblock \url{https://hobbyking.com/}, 2023.

\bibitem{zualkernan2023}
I.~Zualkernan et~al.
\newblock Machine learning for precision agriculture using imagery from
  unmanned aerial vehicles ({UAV}s): A survey.
\newblock {\em Drones}, 7(6):382, Jun 2023.

\end{thebibliography}

\end{document}